\documentstyle[epsf,leqno]{article}

    \makeatletter

\let\@internalcite\cite
\def\cite{\def\citeauthoryear##1##2{##1, ##2}\@internalcite}
\def\shortcite{\def\citeauthoryear##1##2{##2}\@internalcite}
\def\@biblabel#1{\def\citeauthoryear##1##2{##1, ##2}[#1]\hfill}

  \newlength\titlebox 
 \twocolumn 

\def\addcontentsline#1#2#3{}

\def\maketitle{\par
 \begingroup
   \def\thefootnote{\fnsymbol{footnote}}
   \def\@makefnmark{\hbox to 0pt{$^{\@thefnmark}$\hss}}
   \twocolumn[\@maketitle] \@thanks
 \endgroup
 \setcounter{footnote}{0}
 \let\maketitle\relax \let\@maketitle\relax
 \gdef\@thanks{}\gdef\@author{}\gdef\@title{}\let\thanks\relax}
\def\@maketitle{\vbox to \titlebox{\hsize\textwidth
 \linewidth\hsize \vskip 0.625in minus 0.125in \centering
 {\huge\bf \@title \par} \vskip 0.2in plus 1fil minus 0.1in
 {\def\and{\unskip\enspace{\rm and}\enspace}%
  \def\And{\end{tabular}\hss \egroup \hskip 1in plus 2fil 
           \hbox to 0pt\bgroup\hss \begin{tabular}[t]{c}\Large\bf}%
  \def\AND{\end{tabular}\hss\egroup \hfil\hfil\egroup
          \vskip 0.25in plus 1fil minus 0.125in
           \hbox to \linewidth\bgroup\Large \hfil\hfil
             \hbox to 0pt\bgroup\hss \begin{tabular}[t]{c}\Large\bf}
  \hbox to \linewidth\bgroup\Large \hfil\hfil
    \hbox to 0pt\bgroup\hss \begin{tabular}[t]{c}\Large\bf\@author 
                            \end{tabular}\hss\egroup
    \hfil\hfil\egroup}
  \vskip 0.3in plus 2fil minus 0.1in
}}

\def\section{\@startsection {section}{1}{\z@}{-2.0ex plus
    -0.5ex minus -.2ex}{3pt plus 2pt minus 1pt}{\Large\bf\centering}}
\def\subsection{\@startsection{subsection}{2}{\z@}{-2.0ex plus
    -0.5ex minus -.2ex}{3pt plus 2pt minus 1pt}{\large\bf\raggedright}}
\def\subsubsection{\@startsection{subparagraph}{3}{\z@}{-6pt plus
   2pt minus 1pt}{-1em}{\normalsize\bf}}

\skip\footins 9pt plus 4pt minus 2pt
\def\footnoterule{\kern-3pt \hrule width 5pc \kern 2.6pt }
\labelwidth\leftmargini\advance\labelwidth-\labelsep \labelsep 5pt

\def\@listi{\leftmargin\leftmargini}
\def\@listii{\leftmargin\leftmarginii
   \labelwidth\leftmarginii\advance\labelwidth-\labelsep
   \topsep 2pt plus 1pt minus 0.5pt
   \parsep 1pt plus 0.5pt minus 0.5pt
   \itemsep \parsep}
\def\@listiii{\leftmargin\leftmarginiii
    \labelwidth\leftmarginiii\advance\labelwidth-\labelsep
    \topsep 1pt plus 0.5pt minus 0.5pt 
    \parsep \z@ \partopsep 0.5pt plus 0pt minus 0.5pt
    \itemsep \topsep}
\def\@listiv{\leftmargin\leftmarginiv
     \labelwidth\leftmarginiv\advance\labelwidth-\labelsep}
\def\@listv{\leftmargin\leftmarginv
     \labelwidth\leftmarginv\advance\labelwidth-\labelsep}
\def\@listvi{\leftmargin\leftmarginvi
     \labelwidth\leftmarginvi\advance\labelwidth-\labelsep}

\belowdisplayskip \abovedisplayskip
\def\@normalsize{\@setsize\normalsize{11pt}\xpt\@xpt}
\def\small{\@setsize\small{10pt}\ixpt\@ixpt}
\def\footnotesize{\@setsize\footnotesize{10pt}\ixpt\@ixpt}
\def\scriptsize{\@setsize\scriptsize{8pt}\viipt\@viipt}
\def\tiny{\@setsize\tiny{7pt}\vipt\@vipt}
\def\large{\@setsize\large{12pt}\xipt\@xipt}
\def\Large{\@setsize\Large{14pt}\xiipt\@xiipt}
\def\LARGE{\@setsize\LARGE{16pt}\xivpt\@xivpt}
\def\huge{\@setsize\huge{20pt}\xviipt\@xviipt}
\def\Huge{\@setsize\Huge{23pt}\xxpt\@xxpt}

\def\fileversion{1.7}
\def\filedate{20 May 1993}

\newdimen\exampleindent
\divide\exampleindent by 3

\@ifundefined{mathindent}{}{%
  \mathindent=\exampleindent
}

\def\theexample{\theequation}           

\def\examples{%
  \list{]}{%
    \leftmargin=\exampleindent
    \labelwidth=\exampleindent
    \advance\labelwidth -\labelsep
    \def\@listctr{equation}
    \@nmbrlisttrue                      
    \let\makelabel=\@mkexlabel
}}

\def\endexamples{\endlist}

\def\@mkexlabel#1{%
  \if\@itemlabel#1(\theexample)%
  \else\def\@tempa{#1}\ifx\@tempa\@empty{}
  \else (#1)%
  \let\@tempa\protect \def\protect{\noexpand\protect\noexpand}%
  \xdef\@currentlabel{{#1}}%
  \let\protect\@tempa
  \fi\fi\hfil
}

\newcounter{subexample}                 

\def\thesubexample{\alph{subexample}}   

\def\subexamples{%
  \list{]}{%
    \usecounter{subexample}
    \let\makelabel=\@mksubexlabel
}}

\def\endsubexamples{\endlist}

\def\@mksubexlabel#1{%
  \if\@itemlabel#1\thesubexample.%
  \else\def\@tempa{#1}\ifx\@tempa\@empty{}
  \else #1.%
  \let\@tempa\protect \def\protect{\noexpand\protect\noexpand}%
  \xdef\@currentlabel{{#1}}%
  \let\protect\@tempa
  \fi\fi\hfil
}

\newcounter{subsubexample}

\def\thesubsubexample{\roman{subsubexample}}

\def\subsubexamples{%
  \list{]}{%
    \usecounter{subsubexample}
    \let\makelabel=\@mksubsubexlabel
}}

\def\@mksubsubexlabel#1{%
  \if\@itemlabel#1\thesubsubexample.%
  \else\def\@tempa{#1}\ifx\@tempa\@empty{}
  \else #1.%
  \let\@tempa\protect \def\protect{\noexpand\protect\noexpand}%
  \xdef\@currentlabel{{#1}}%
  \let\protect\@tempa
  \fi\fi\hfil
}

\def\subex{\examples \item \bgroup \subexamples}
\def\endsubex{\endsubexamples \egroup \endexamples}

\def\sqz#1{\leavevmode \llap{#1\kern 0.1 em}}

\def\attop#1{%
  \leavevmode
  \vtop{\vbox to \baselineskip{}\kern -\baselineskip \hbox{#1}}%
}

   \makeatother

\newcommand{\savm}[1]{\mbox{\tiny \(
                             \setlength{\arraycolsep}{.4ex}
                             \renewcommand{\arraystretch}{1.0} 
                             \hspace*{-1.4ex}\left[ 
                             \begin{array}{ll}  
                             \\[-1ex] #1 \\[-1.4ex]
                             \end{array}
                             \right]\hspace*{-2.3ex}
                           \)
                    }}

\newcommand{\attval}[2]{\hspace*{-1mm}
                        \mbox{\uppercase{#1}}
                        & 
                        #2 \hspace*{-.25ex}\\}

\newcommand{\set}[1]{\mbox{\(
                             \setlength{\arraycolsep}{.4ex}
                             \renewcommand{\arraystretch}{1.1}      
                             \hspace*{-1.0ex}\left\{
                              \begin{array}{l}
                               #1
                               \end{array}
                             \right\}\hspace*{-1.5ex}
                           \)
                    }}

\newcommand{\pbar}{\mbox{$\mid$}}

\newcommand{\ind}[1]{\mbox{${\setlength{\fboxsep}{0.5mm}
        \fbox{{\tiny #1}} }$}}

    \def\bfitem[#1]{\def\protect##1{\let\protect=\relax}%
      \item[\protect\noexpand\bf #1]\protect\relax}

\newcommand{\lscott}{[ \hspace{-.5mm} [}
\newcommand{\rscott}{] \hspace{-.5mm} ]}
\newcommand{\denotation}[1]{\lscott #1 \rscott}

\newcommand{\nfrac}[2]{$\frac{\mbox{#1}}{\mbox{#2}}$}

\newtheorem{theorem}{Theorem}
\newtheorem{lemma}[theorem]{Lemma}

\newcommand{\C}{\mbox{${\cal C}$}}

\newcommand{\F}{\mbox{${\cal F}$}}

\newcommand{\I}{{\cal I}}

\newcommand{\PP}{\mbox{${\cal P}$}}
\newcommand{\R}{{\cal R}}

\newcommand{\U}{\mbox{${\cal U}$}}
\newcommand{\V}{{\cal V}}

\newcommand{\At}{\mbox{$\cal A$}t}

\newcommand{\card}[1]{\hspace{.2em}\mid \! #1 \!\mid\hspace{.2em}}

\title{\vspace{-0.5in}An Attributive Logic of Set Descriptions and\\ Set Operations}

\author{Suresh Manandhar\\
HCRC Language Technology Group \\
{\em The University of Edinburgh} \\
2 Buccleuch Place\\
Edinburgh EH8 9LW, UK\\
Internet: {\tt Suresh.Manandhar@ed.ac.uk}
}

\begin{document}

\maketitle
\vspace{1ex}
\begin{abstract}

This paper provides a model theoretic semantics to feature
  terms augmented with set descriptions.  We provide constraints to
  specify HPSG style set descriptions, fixed cardinality set
  descriptions, set-membership constraints, restricted universal role
  quantifications, set union, intersection, subset and disjointness.  A
  sound, complete and terminating consistency checking procedure is
  provided to determine the consistency of any given term in the logic.
  It is shown that determining consistency of terms is a NP-complete problem.
\end{abstract}

{\bf Subject Areas:}  feature logic, constraint-based grammars, HPSG

\section{Introduction}
Grammatical formalisms such as HPSG \cite{Pollard:hpsg1}
\cite{Pollard:hpsg2} and LFG \cite{kb82} employ feature descriptions
\cite{Kasper:logical} \cite{Smolka:constraint} as the primary means
for stating linguistic theories. However the descriptive machinery
employed by these formalisms easily exceed the descriptive machinery
available in feature logic \cite{Smolka:constraint}. Furthermore the
descriptive machinery employed by both HPSG and LFG is difficult (if
not impossible) to state in feature based formalisms such as
ALE~\cite{Carpenter:ALE}, TFS~\cite{Zajac:Inheritance} and
CUF~\cite{DorreDorna:CUF} which augment feature logic with a type
system. One such expressive device employed both within
LFG~\cite{kb82} and HPSG but is unavailable in feature logic is that
of set descriptions.

Although various researchers have studied set descriptions (with
different semantics) \cite{Rounds:values} \cite{Pollard:unifying} two
issues remain unaddressed. Firstly there has not been any work on
consistency checking techniques for feature terms augmented with set
descriptions.  Secondly, for applications within grammatical theories
such as the HPSG formalism, set descriptions alone are not enough
since descriptions involving set union are also needed. Thus to
adequately address the knowledge representation needs of current
linguistic theories one needs to provide set descriptions as well as
mechanisms to manipulate these.

In the HPSG grammar formalism \cite{Pollard:hpsg1}, set descriptions are
employed for the modelling of so called {\sl semantic indices}
(\cite{Pollard:hpsg1} {\sl pp.} 104). The attribute {\sc inds} in the
example in (\ref{ex:SemInd}) is a multi-valued attribute whose value
models a set consisting of (at most) 2 objects. However multi-valued
attributes cannot be described within feature logic \cite{Kasper:logical}
\cite{Smolka:constraint}.

\newsavebox{\bigstruct}
\sbox{\bigstruct}{$
\set{\savm{\attval{var}{\ind{1}}
                               \attval{rest}{\savm{\attval{reln}{naming}
                                                  \attval{name}{sandy}
                                                  \attval{named}{\ind{1}}}}},
                          \savm{\attval{var}{\ind{2}}
                               \attval{rest}{\savm{\attval{reln}{naming}
                                                  \attval{name}{kim}
                                                  \attval{named}{\ind{2}}}}}}
$}
\begin{ex}                                   \label{ex:SemInd}
\ \\
\hspace*{-4ex}\attop{
$ \savm{\attval{cont}{\savm{\attval{rel}{see}
                          \attval{seer}{\ind{2}}
                          \attval{seen}{\ind{1}}}}
       \attval{inds}{\usebox{\bigstruct}
                     }}
$
}
\end{ex}

A further complication arises since to be able to deal with anaphoric
dependencies we think that set memberships will be needed to resolve
pronoun dependencies. Equally, set unions may be called for to
incrementally construct discourse referents. Thus set-valued extension
to feature logic is insufficient on its own.

Similarly, set valued subcategorisation frames (see (\ref{ex:bsubcat}))
has been considered as a possibility within the HPSG formalism.
\begin{ex}                                 \label{ex:bsubcat}
\ \\
\hspace*{-4ex}
\attop{
$believes = 
\savm{
  \attval{syn{\pbar}loc{\pbar}subcat}{
    \set{
      \savm{
        \attval{syn{\pbar}loc{\pbar}head{\pbar}cat}{n}}\ ,\\
      \savm{
        \attval{syn{\pbar}loc{\pbar}head{\pbar}cat}{v}}
      }
    }}$
}
\end{ex}
But once set valued subcategorisation frames are employed, a set
valued analog of the HPSG subcategorisation principle too is needed.
In section \ref{sec:SetLogic} we show that the set valued analog of
the subcategorisation principle can be adequately described by
employing a disjoint union operation over set descriptions as
available within the logic described in this paper.

\section{The logic of Set descriptions}
\label{sec:SetLogic}
In this section we provide the semantics of feature terms augmented with
set descriptions and various constraints over set descriptions.  We
assume an alphabet consisting of $x, y, z, \ldots \in \V$ the set of {\em
  variables}; $f, g, \ldots \in \F$ the set of {\em relation symbols};
$c_{1}, c_{2}, \ldots \in \C$ the set of {\em constant symbols}; $A, B,
C, \ldots \in \PP$ the set of primitive concept symbols and $a, b, \ldots
\in \At$ the set of {\em atomic symbols}. Furthermore, we require that
$\bot, \top \in \PP$.

The syntax of our term language defined by the following BNF definition:
\begin{enumerate}
\item[] \hspace*{-2.6ex}$P \longrightarrow 
                       x 
                       \mid a 
                       \mid c 
                       \mid C 
                       \mid \neg x 
                       \mid \neg a 
                       \mid \neg c 
                       \mid \neg C$

\item[] \hspace*{-2.8ex}$\setlength{\arraycolsep}{.2ex}
         \begin{array}{llll}
         S, T & \longrightarrow \\
         & P \\
         & \mid f:T          & \mbox{\rm feature term}\\
         & \mid \exists f:T  & \mbox{\rm existential role quantification}\\
         & \mid \forall f:P  & \mbox{\rm universal role quantification}\\
         & \mid f:\{T_{1}, \ldots, T_{n} \} & \mbox{\rm set description}\\
         & \mid f:\{T_{1}, \ldots, T_{n}\}_{=} & \mbox{\rm fixed
           cardinality set description}\\
         & \mid f:g(x) \cup h(y) & \mbox{\rm union}\\
         & \mid f:g(x) \cap h(y) & \mbox{\rm intersection}\\
         & \mid f:\supseteq g(x) & \mbox{\rm subset}\\
         & \mid f(x) \neq g(y) & \mbox{\rm disjointness}\\
         & \mid S \sqcap T   & \mbox{\rm  conjunction}
         \end{array}$
\end{enumerate}
where $S, T, T_{1}, \ldots, T_{n}$ are terms; $a$ is an
{\em atom}; $c$ is a {\em constant}; $C$ is a {\em primitive concept} and
$f$ is a {\em relation symbol}.

The interpretation of {\em relation symbols} and {\em atoms} is provided
by an interpretation $\I = <\U^{I}, I>$ where $\U^{I}$ is an arbitrary
non-empty set and $I$ is an interpretation function that maps :
\begin{enumerate}
\item every relation symbol $f \in \F$ to a binary relation $f^{I}
  \subseteq \U^{I} \times \U^{I}$

\item every atom $a \in \At$ to an element $a^{I} \in \U^{I}$
\end{enumerate}
{\bf Notation:}
\begin{itemize}
\item Let $f^{I}(e)$ denote the set $\{ e' \mid (e, e')
  \in f^{I} \}$
\item Let $f^{I}(e) \uparrow$ mean $f^{I}(e) = \emptyset$
\end{itemize}
$\I$ is required to satisfy the following properties :
      \begin{enumerate}
      \item if $a_{1} \not\equiv a_{2}$ then $a_{1}^{I} \neq a_{2}^{I}$\
        \ \ \ \ \ ({\em distinctness})
      \item for any atom $a \in \At$ and for any relation $f \in \F$
        there exists no $e \in \U^{I}$ such that
            $(a, e) \in f^{I}$  \ \ \ \ \ ({\em atomicity})
\end{enumerate}
\noindent For a given interpretation $\I$ an {\bf $\I$-assignment}
$\alpha$ is a function that maps :
\begin{enumerate}
\item every variable $x \in \V$ to an element $\alpha(x) \in \U^{I}$
\item every constant $c \in \C$ to an element $\alpha(c) \in \U^{I}$ such
  that for distinct constants $c_{1}, c_{2}$ : 
  $\alpha(c_{1}) \neq \alpha(c_{2})$
\item every primitive concept $C \in \PP$ to a subset 
$\alpha(C) \subseteq \U^{I}$ such that:
 \begin{itemize}
 \item $\alpha(\bot) = \emptyset$
 \item $\alpha(\top) = \U^{I}$
 \end{itemize}
\end{enumerate}
\noindent The interpretation of  terms is provided by a denotation
function $\denotation{.}^{\I,\alpha}$ that given an interpretation
$\I$ and an $\I$-assignment $\alpha$ maps terms to subsets of $\U^{I}$.

The  function $\denotation{.}^{\I,\alpha}$ is defined as
follows :
\begin{enumerate}
\item[] $\denotation{x}^{\I,\alpha} = \{ \alpha(x) \}$
\item[] $\denotation{a}^{\I,\alpha} = \{ a^{I} \}$
\item[] $\denotation{c}^{\I,\alpha} = \{ \alpha(c) \}$
\item[] $\denotation{C}^{\I,\alpha} = \alpha(C)$
\item[] $\denotation{f:T}^{\I,\alpha} =$\\
        \hspace*{2ex}$\{ e \in \U^{I} \mid \exists e' \in \U^{I} :
                         f^{I}(e) = \{ e' \} \ \wedge \ 
                        e' \in \denotation{T}^{\I,\alpha} \}$

\item[] $\denotation{\exists f:T}^{\I,\alpha} =$\\
        \hspace*{2ex}$\{ e \in \U^{I} \mid \exists e' \in \U^{I} :
                        (e, e') \in f^{I}\ \wedge \ 
                        e' \in \denotation{T}^{\I,\alpha} \}$

\item[] $\denotation{\forall f:T}^{\I,\alpha} =$\\
        \hspace*{2ex}$\{ e \in \U^{I} \mid \forall e' \in \U^{I} :
                         (e, e') \in f^{I} \Rightarrow 
                        e' \in \denotation{T}^{\I,\alpha} \}$

\item[] $\denotation{f:\{T_{1},\ldots, T_{n} \}}^{\I,\alpha} =$\\
        \hspace*{2ex}$\{ e \in \U^{I} \mid 
                       \exists e_{1}, \ldots, \exists e_{n} \in \U^{I}:$\\
           \hspace*{7ex}$f^{I}(e) = \{e_{1}, \ldots, e_{n}\} \wedge$\\
           \hspace*{7ex}$e_{1} \in \denotation{T_{1}}^{\I,\alpha} 
                        \wedge \ldots \wedge
                        e_{n} \in \denotation{T_{n}}^{\I,\alpha}\}$

\item[] $\denotation{f:\{T_{1},\ldots, T_{n} \}_{=}}^{\I,\alpha} =$\\
      \hspace*{2ex}$\{ e \in \U^{I} \mid
              \exists e_{1}, \ldots, \exists e_{n} \in \U^{I}:$\\
      \hspace*{7ex}$\card{f^{I}(e)} = n \wedge 
                    f^{I}(e) = \{e_{1}, \ldots, e_{n}\} \wedge$\\
      \hspace*{7ex}$e_{1} \in \denotation{T_{1}}^{\I,\alpha} 
                        \wedge \ldots \wedge
                        e_{n} \in \denotation{T_{n}}^{\I,\alpha}\}$

\item[] $\denotation{f:g(x) \cup h(y)}^{\I,\alpha} =$\\
        \hspace*{2ex}$\{ e \in \U^{I} \mid 
                 f^{I}(e) = g^{I}(\alpha(x)) \cup h^{I}(\alpha(y))
            \}$
\item[] $\denotation{f:g(x) \cap h(y)}^{\I,\alpha} =$\\
        \hspace*{2ex}$\{ e \in \U^{I} \mid 
                 f^{I}(e) = g^{I}(\alpha(x)) \cap h^{I}(\alpha(y))
            \}$
\item[] $\denotation{f: \supseteq g(x)}^{\I,\alpha} =$\\
        \hspace*{2ex}$\{ e \in \U^{I} \mid 
                 f^{I}(e) \supseteq g^{I}(\alpha(x))
            \}$
\item[] $\denotation{f(x) \neq g(y)}^{\I,\alpha} =$
        \hspace*{2ex}\begin{itemize}
         \item $\emptyset$ if 
            $f^{I}(\alpha(x)) \cap g^{I}(\alpha(y)) \neq \emptyset$
         \item $\U^{I}$ if $f^{I}(\alpha(x)) \cap g^{I}(\alpha(y)) = \emptyset$
         \end{itemize}
\item[] $\denotation{S \sqcap T}^{\I,\alpha} = 
                  \denotation{S}^{\I,\alpha} \cap  
                  \denotation{T}^{\I,\alpha}$

\item[] $\denotation{\neg T}^{\I,\alpha} = 
                      \U^{I} - \denotation{T}^{\I,\alpha}$

\end{enumerate}
\noindent The above definitions fix the syntax and semantics of every term.

\noindent It follows from the above definitions that:\\
$f:T~\equiv~f:\{T\}~\equiv f:\{T\}_{=}$

\begin{figure}[htb]
\begin{center} 
\mbox{\epsffile{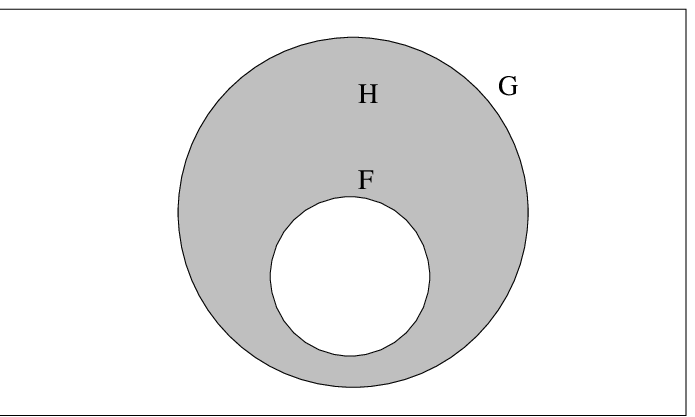}}
\end{center}
\caption{ }
\label{fig:venn}
\vspace*{-2ex}
\end{figure}

Although {\em disjoint union} is not a primitive in the logic it can
easily be defined by employing set disjointness and set union operations:
\begin{itemize}
\item[] $f: g(x) \uplus h(y) =_{def}\ \  
         g(x) \neq h(y) \sqcap f:g(x) \cup h(y)$
\end{itemize}
Thus disjoint set union is exactly like set union except that it
additionally requires the sets denoted by $g(x)$ and $h(y)$ to be
disjoint.

\begin{figure*}[htb]
\begin{center} 
\framebox[40em][l]{
\parbox[t]{40em}{
\begin{center} 
{\large \bf Decomposition rules}
\end{center}
\begin{examples}

\bfitem[DFeat]                                    \label{rule:Dfeat}
  \nfrac{$x = F : T \wedge C_{s}$}{  
              $x = F : y \wedge y = T \wedge  C_{s}$}\\
       if $y$ is new and $T$ is not a variable and $F$ ranges over
       $\exists f, f$


\bfitem[DForall]                                     \label{rule:DForall}
 \nfrac{$x = \forall f: \overline{c} \wedge C_{s}$}{
        $x = \forall f: y \wedge y = \overline{c} \wedge C_{s}$}\\
       if $y$ is new and $\overline{c}$ ranges over $a, c$.

\bfitem[DSet]                                     \label{rule:DSet}
  \nfrac{$x = f:\{T_{1},\ldots,T_{n}\} \wedge C_{s}$}{
       $x = f:\{x_{1},\ldots,x_{n}\} \wedge 
        x_{1} = T_{1} \wedge \ldots \wedge x_{n} = T_{n} \wedge C_{s}$}\\
       if $x_{1}, \ldots, x_{n}$ are new
       and at least one of $T_{i} : 1 \leq i \leq n$ is not a variable 

\bfitem[DSetF]                                     \label{rule:DSetF}
  \nfrac{$x = f:\{T_{1},\ldots,T_{n}\}_{=} \wedge C_{s}$}{
       $x = f:\{x_{1},\ldots,x_{n}\} \wedge
        x = f:\{x_{1},\ldots,x_{n}\}_{=} \wedge
        x_{1} = T_{1} \wedge \ldots \wedge x_{n} = T_{n} \wedge C_{s}$}\\
       if $x_{1}, \ldots, x_{n}$ are new
       and at least one of $T_{i} : 1 \leq i \leq n$ is not a variable

\bfitem[DConj]                                     \label{rule:DConj}
 \nfrac{$x =  S \sqcap T \wedge  C_{s}$}{
        $x =  S \wedge  
         x =  T \wedge  C_{s}$}

\end{examples}
}}
\end{center}
\caption{Decomposition rules}
\label{fig:DRules}
\vspace*{-2ex}
\end{figure*}

The set-valued description of the subcategorisation principle can now be
stated as given in example (\ref{ex:SetSubcat}).
\begin{ex}                               \label{ex:SetSubcat}
\hspace*{-1ex}{\bf Subcategorisation Principle}\\
\hspace*{-3ex}$\savm{\attval{syn{\pbar}loc}{Y}
        \attval{dtrs}{X \sqcap\ 
          \savm{\attval{h-dtr{\pbar}syn{\pbar}loc{\pbar}subcat}
                        {\mbox{c-dtrs}(X) \uplus \mbox{subcat}(Y)}
               }
             }
           }$
\end{ex}
The description in (\ref{ex:SetSubcat}) simply states that the subcat
value of the \mbox{\sc h-dtr} is the disjoint union of the subcat
value of the mother and the values of \mbox{\sc c-dtrs}. Note that
the disjoint union operation is the right operation to be specified to split
the set into two disjoint subsets.  Employing just union operation would
not work since it would permit repetition between members of the
\mbox{\sc subcat} attribute and \mbox{\sc c-dtrs} attribute.

Alternatively, we can assume that {\sc n} is the only multi-valued
relation symbol while both {\sc subcat} and {\sc c-dtrs} are
single-valued and then employ the intuitively appealing subcategorisation
principle given in (\ref{ex:SetSubcat2}).  
\begin{ex}                               \label{ex:SetSubcat2}
\hspace*{-1ex}{\bf Subcategorisation Principle}\\
\hspace*{-1ex}$\savm{\attval{syn{\pbar}loc{\pbar}subcat}{Y}
        \attval{dtrs}{
          \savm{\attval{h-dtr{\pbar}syn{\pbar}loc{\pbar}subcat{\pbar}n}
                        {\mbox{N}(X) \uplus \mbox{N}(Y)}
                \attval{c-dtrs}{X}
               }
             }
           }$
\end{ex}
With the availability of set operations, multi-valued structures can
be incrementally built. For instance, by employing union operations,
semantic indices can be incrementally constructed and by employing
membership constraints on the set of semantic indices pronoun resolution
may be carried out.

The set difference operation $f:g(y) - h(z)$ is not available from the
constructs described so far. However, assume that we are given the term
$x \sqcap f:g(y) - h(z)$ and it is known that $h^{\I}(\alpha(z))
\subseteq g^{\I}(\alpha(y))$ for every interpretation $\I, \alpha$ such
that $\denotation{x \sqcap f:g(y) - h(z)}^{\I,\alpha} \neq \emptyset$.
Then the term $x \sqcap f:g(y) - h(z)$ (assuming the obvious
interpretation for the set difference operation) is consistent iff the
term $y \sqcap g: f(x) \uplus h(z)$ is consistent. This is so since for
sets G, F, H :\ G - F = H $\wedge$ F $\subseteq$ G {\em iff} G = F
$\uplus$ H. See figure \ref{fig:venn} for verification.

\section{Consistency checking}
To employ a term language for knowledge representation tasks or in
constraint programming languages the minimal operation that needs to be
supported is that of consistency checking of terms.

\begin{figure}[htb]
\begin{center} 
\framebox[\linewidth][l]{
\parbox[t]{\linewidth}{
\begin{center} 
{\large \bf Constraint simplification rules - I}
\end{center}
\begin{examples}
\bfitem[SEquals]                                  \label{rule:SEquals}
     \nfrac{$x = y \wedge C_{s}$}{ 
        $x = y \wedge [x/y]C_{s}$}\\
       if $x \not\equiv y$ and $x$ occurs in $C_{s}$

\bfitem[SConst]                                  \label{rule:SConst}
   \nfrac{$x = \overline{c} \wedge y  =  \overline{c}  \wedge  C_{s}$}{ 
        $x = y \wedge x  = \overline{c}  \wedge C_{s}$}\\
        where $\overline{c}$ ranges over $a, c$.

\bfitem[SFeat]                                  \label{rule:SFeat}
  \nfrac{$x = f: y \wedge  x = F : z   \wedge  C_{s}$}{ 
        $x = f: y \wedge y = z  \wedge  C_{s} $}\\
        where $F$ ranges over $f, \exists f, \forall f$

\bfitem[SExists]                                  \label{rule:SExists}
        \nfrac{$x = \exists f: y \wedge
          x = \forall f: z \wedge C_{s}$}{
         $x = f: y \wedge y = z  \wedge C_{s}$}

\bfitem[SForallE]
        \nfrac{$x = \forall f: \overline{C} \wedge
          x = \exists f : y  \wedge C_{s}$}{
         $x = \forall f: \overline{C} \wedge x = \exists f: y \wedge
          y = \overline{C}  \wedge C_{s}$}\\
        if $\overline{C}$ ranges over $C, \neg C, \neg a, \neg c, \neg z$ and\\
           $C_{s} \not\vdash y = \overline{C}$.

\end{examples}
}
}
\end{center}
\caption{Constraint simplification rules - I}
\label{fig:SRules1}
\vspace*{-3ex}
\end{figure}

A term $T$ is {\bf consistent} if there exists an interpretation $\I$
and an $\I$-assignment $\alpha$ such that
$\denotation{T}^{\I,\alpha}\neq\emptyset$.

In order to develop constraint solving algorithms for consistency testing
of terms we follow the approaches in \cite{Smolka:constraint}
\cite{Hollunder:subsumption}.

\begin{figure*}[htb]
\begin{center} 
\framebox[30em][l]{
\parbox[t]{30em}{
\begin{center} 
{\large \bf Constraint simplification rules - II}
\end{center}
\begin{examples}
\bfitem[SSetF]                              \label{rule:SSetF}
 \nfrac{$ x = F : y \wedge x = f:\{x_{1}, \ldots, x_{n} \} \wedge C_{s}$}{
       $x = f : y \wedge y =  x_{1} \wedge \ldots \wedge y = x_{n} 
       \wedge C_{s}$}\\
       where $F$ ranges over $f, \forall f$

\bfitem[SSet]                                      \label{rule:SSet}
 \nfrac{$ x = f:\{y\}  \wedge C_{s}$}{
       $x = f: y  \wedge C_{s}$}

\bfitem[SDup]                                      \label{rule:SDup}
     \nfrac{$x = f:\{x_{1},\ldots, x_{i},\ldots,x_{j},\ldots,x_{n}\} 
       \wedge C_{s}$}{
       $x = f:\{x_{1},\ldots, x_{i},\ldots, \ldots,x_{n}\} \wedge C_{s}$}\\
       if $x_{i} \equiv x_{j}$

\bfitem[SForall]                                      \label{rule:SForall}
        \nfrac{$x = \forall f : \overline{C} \wedge
         \ x = f: \{ x_{1}, \ldots, x_{n} \} \wedge C_{s}$}{
        $x = f: \{ x_{1}, \ldots, x_{n} \} \wedge
          x_{1} = \overline{C} \wedge \ldots \wedge x_{n} = \overline{C}
        \wedge C_{s}$}\\
       if $\overline{C}$ ranges over $C, \neg C, \neg a, \neg c, \neg z$ and\\ 
        there exists $x_{i} : 1 \leq i \leq n$ such that 
        $C_{s} \not\vdash x_{i} = \overline{C}$.

\bfitem[SSetE]
 \nfrac{$x = \exists f: y \wedge x = f:\{x_{1}, \ldots, x_{n} \}  
               \wedge C_{s}$}{
         $x = f:\{x_{1}, \ldots, x_{n} \} \wedge
          y = x_{1} \sqcup \ldots \sqcup x_{n}  \wedge C_{s}$}

\bfitem[SSetSet]                                         \label{rule:SSetSet}
      \nfrac{$ x =  f:\{x_{1}, \ldots, x_{n} \} \wedge
          x = f:\{y_{1}, \ldots, y_{m} \} \wedge C_{s}$}{
        $\begin{array}{lll}
           x = f:\{x_{1}, \ldots, x_{n} \} \wedge \\
          x_{1} = y_{1} \sqcup  \ldots \sqcup y_{m} \wedge 
          \ldots \wedge 
          x_{n} = y_{1} \sqcup  \ldots \sqcup y_{m} \wedge \\
          y_{1} = x_{1} \sqcup  \ldots \sqcup x_{n} \wedge
          \ldots \wedge 
          y_{m} = x_{1} \sqcup  \ldots \sqcup x_{n}  \wedge C_{s}
         \end{array}$}  \\
          where $n \leq m$

\bfitem[SDis]                                          \label{rule:SDis}
       \nfrac{$ x = x_{1} \sqcup \ldots \sqcup x_{n}  \wedge C_{s}$}{
         $x = x_{1} \sqcup \ldots \sqcup x_{n} \wedge
          x = x_{i}  \wedge  C_{s}$}\\
         if $1 \leq i \leq n$ and\\
         there is no $x_{j}, 1 \leq j \leq n$ such that
         $C_{s} \vdash x = x_{j}$

\end{examples}
}
}
\caption{Constraint simplification rules - II}
\label{fig:SRules2}
\end{center}
\vspace*{-3ex}
\end{figure*}

A {\bf containment constraint} is a constraint of the form $x = T$
where $x$ is a variable and $T$ is an  term.

In addition, for the purposes of consistency checking we need to introduce
{\bf disjunctive constraints} which are of the form \mbox{$x= x_{1}\sqcup
\ldots \sqcup x_{n}$}.

We say that an interpretation $\I$ and an $\I$-assignment $\alpha$
satisfies a constraint $K$ written $\I, \alpha \models K$ {\em if}:
\begin{itemize}
\item $\I, \alpha \models x = T \Longleftrightarrow
                          \alpha(x) \in \denotation{T}^{\I,\alpha}$
\item $\I, \alpha \models x =  x_{1} \sqcup \ldots \sqcup x_{n}
          \Longleftrightarrow
          \alpha(x) = \alpha(x_{i})$ for some $x_{i} : 1 \leq i \leq n$.
\end{itemize}

A {\bf constraint system} $C_{s}$ is a {\em conjunction} of constraints.

We say that an interpretation $\I$ and an $\I$-assignment $\alpha$
{\bf satisfy} a constraint system $C_{s}$ {\em iff} $\I, \alpha$
satisfies every constraint in $C_{s}$.

The following lemma demonstrates the usefulness of constraint systems for
the purposes of consistency checking.
\begin{lemma}
  An term $T$ is consistent iff there exists a variable $x$, an
  interpretation $\I$ and an $\I$-assignment $\alpha$ such
  that $\I, \alpha$ satisfies the constraint system $x = T$.
\end{lemma}
\noindent Now we are ready to turn our attention to constraint solving
rules that will allow us to determine the consistency of a given
constraint system.

We say that a constraint system $C_{s}$ is {\bf basic} if {\em none} of
the {\em decomposition rules} (see figure \ref{fig:DRules}) are
applicable to $C_{s}$.

The purpose of the decomposition rules is to break down a complex
constraint into possibly a number of simpler constraints upon which the
constraint simplification rules (see figures \ref{fig:SRules1},
\ref{fig:SRules2} and 5
) can apply by possibly introducing new variables.

The first phase of consistency checking of a term $T$ consists of
exhaustively applying the decomposition rules to an initial constraint of
the form $x = T$ (where $x$ does not occur in $T$) until no rules
are applicable. This transforms any given constraint system into {\sl
  basic form}.

The constraint simplification rules (see figures \ref{fig:SRules1},
\ref{fig:SRules2} and 5
) either eliminate variable equalities of the form $x=y$ or generate them
from existing constraints.  However, they do not introduce new variables.

The constraint simplification rules given in figure \ref{fig:SRules1} are
the analog of the feature simplification rules provided in
\cite{Smolka:logic}. The main difference being that our simplification
rules have been modified to deal with relation symbols as opposed to just
feature symbols.

The constraint simplification rules given in figure \ref{fig:SRules2}
simplify constraints involving set descriptions when they interact with
other constraints such as feature constraints - rule (\ref{rule:SSetF}),
singleton sets - rule (\ref{rule:SSet}), duplicate elements in a set -
rule (\ref{rule:SDup}), universally quantified constraint - rule
(\ref{rule:SForall}), another set description - rule
(\ref{rule:SSetSet}). Rule (\ref{rule:SDis}) on the other hand simplifies
disjunctive constraints. Amongst all the constraint simplification rules
in figures \ref{fig:SRules1} and \ref{fig:SRules2} only rule
(\ref{rule:SDis}) is non-deterministic and creates a $n$-ary choice
point.

\begin{figure}[htb]
\begin{center} 
\framebox[\linewidth][l]{
\parbox[t]{\linewidth}{
{\large \bf  Extended Constraint simplification rules}
\begin{examples}
\bfitem[$\subseteq$]                               \label{rule:Subset}
     \nfrac{$x = f: \supseteq g(y)  \wedge C_{s}$}{
       $x = f: \supseteq g(y) \wedge
        x = \exists f: y_{i} \wedge C_{s}$}\\
       if:
        \begin{itemize}
        \item $C_{s} \not\vdash  x = \exists f: y_{i}$ and
        \item $C_{s} \vdash  y = \exists g: y_{i}$
        \end{itemize}

\bfitem[$\cup Left$]                           \label{rule:UnionLeft}
     \nfrac{$x = f: g(y) \cup h(z)\wedge C_{s}$}{
       $x = f: g(y) \cup h(z) \wedge
        x = f: \supseteq g(y) \wedge  C_{s}$}\\
       if  $C_{s} \not\vdash x = f: \supseteq g(y)$ 

\bfitem[$\cup Right$]                          \label{rule:UnionRight}
     \nfrac{$x = f: g(y) \cup h(z)\wedge C_{s}$}{
       $x = f: g(y) \cup h(z) \wedge
        x = f: \supseteq h(z) \wedge  C_{s}$}\\
       if  $C_{s} \not\vdash x = f: \supseteq h(z)$

\bfitem[$\cup Down$]                                 \label{rule:UnionDown}
\ \\
{\ \hspace*{-5ex}}
{\nfrac{$x = f: g(y)$ $\cup$ $h(z)  \wedge C_{s}$}{
       $x = f: g(y) \cup h(z)  \wedge
        y = \exists g: x_{i}\mid z = \exists h: x_{i}\wedge C_{s}$}
      }\\

       if:
        \begin{itemize}
        \item $C_{s} \not\vdash  y = \exists g: x_{i}$ and
        \item $C_{s} \not\vdash  z = \exists h: x_{i}$ and
        \item $C_{s} \vdash  x = \exists f: x_{i}$
        \end{itemize}

\bfitem[$\cap Down$]                      \label{rule:IntersectionDown}
\ \\
{\ \hspace*{-5ex}}
{\nfrac{$x = f: g(y) \cap h(z) \wedge C_{s}$}{
       $x = f: g(y) \cap h(z) \wedge
        y = \exists g: x_{i} \wedge 
        z = \exists h: x_{i} \wedge C_{s}$}
    }\\
   \ \\
    \hspace*{-2ex}\parbox[t]{25em}{if:
        \begin{itemize}
        \item \hspace*{-1ex}($C_{s} \not\vdash  y = \exists g: x_{i}$ or
            $C_{s} \not\vdash  z = \exists h: x_{i}$) and
        \item $C_{s} \vdash  x = \exists f: x_{i}$
        \end{itemize}
    }

\bfitem[$\cap Up$]                        \label{rule:IntersectionUp}
     \nfrac{$x = f: g(y) \cap h(z) \wedge  C_{s}$}{
       $x = f: g(y) \cap h(z) \wedge
        x = \exists f: x_{i} \wedge C_{s}$}\\
       if:
        \begin{itemize}
        \item $C_{s} \not\vdash  x = \exists f: x_{i}$ and
        \item $C_{s} \vdash  y = \exists g: x_{i}$ and
        \item $C_{s} \vdash  z = \exists h: x_{i}$
        \end{itemize}
\end{examples}
}
}
\caption{Constraint solving with set operations}
\label{fig:SRules3}
\end{center}
\vspace*{-3ex}
\end{figure}

Rules (\ref{rule:SSet}) and (\ref{rule:SDup}) are redundant as
completeness (see section below) is not affected by
these rules. However these rules result in a simpler normal form.

The following syntactic notion of entailment is employed to render a
slightly compact presentation of the constraint solving rules for dealing
with set operations given in figure 
5.

A constraint system $C_{s}$ {\em syntactically entails} the (conjunction
of) constraint(s) $\phi$ if $C_{s} \vdash \phi$ is derivable from the
following deduction rules:
\begin{enumerate}
\item $\phi \wedge C_{s} \vdash \phi$
\item $C_{s} \vdash x = x$
\item $C_{s} \vdash x = y \longrightarrow C_{s} \vdash y = x$
\item $C_{s} \vdash x = y \wedge C_{s} \vdash y = z
  \longrightarrow C_{s} \vdash x = z$
\item $C_{s} \vdash x = \neg y \longrightarrow C_{s} \vdash y = \neg x$
\item $C_{s} \vdash x =  f: y \longrightarrow 
       C_{s} \vdash x = \exists f: y$
\item $C_{s} \vdash x =  f: y \longrightarrow 
       C_{s} \vdash x = \forall f: y$
\item $C_{s} \vdash x =  f: \{ \ldots, x_{i}, \ldots \} \longrightarrow 
       C_{s} \vdash x = \exists f: x_{i}$
\end{enumerate}
Note that the above definitions are an incomplete list of deduction rules.
However  $C_{s} \vdash \phi$ implies $C_{s} \models \phi$ where
$\models$ is the semantic entailment relation defined as for predicate
logic.

We write $C_{s} \not\vdash \phi$ if it is not the case that $C_{s} \vdash
\phi$.

The constraint simplification rules given in figure 
5 deal with constraints involving set operations. Rule
(\ref{rule:Subset}) propagates $g$-values of $y$ into $f$-values of $x$
in the presence of the constraint $x = f: \supseteq g(y)$. Rule
(\ref{rule:UnionLeft}) (correspondingly Rule (\ref{rule:UnionRight}))
adds the constraint $x = f: \supseteq g(y)$ (correspondingly $x = f:
\supseteq h(z)$) in the presence of the constraint $x = f: g(y) \cup
h(z)$. Also in the presence of $x = f: g(y) \cup h(z)$ rule
(\ref{rule:UnionDown}) non-deterministically propagates an $f$-value of
$x$ to either an $g$-value of $y$ or an $h$-value of $z$ (if neither
already holds). The notation $y = \exists g: x_{i}\mid z = \exists h:
x_{i}$ denotes a non-deterministic choice between $y = \exists g: x_{i}$
and $z = \exists h: x_{i}$. Rule (\ref{rule:IntersectionDown}) propagates
an $f$-value of $x$ both as a $g$-value of $y$ and $h$-value of $z$ in
the presence of the constraint $x = f: g(y) \cap h(z)$. Finally, rule
(\ref{rule:IntersectionUp}) propagates a common $g$-value of $y$ and
$h$-value of $z$ as an $f$-value of $x$ in the presence of the constraint
$x = f: g(y) \cap h(z)$.

\section{Invariance, Completeness and Termination}
\label{sec:completeness}
In this section we establish the main results of this paper - namely
that our consistency checking procedure for set descriptions and set
operations is invariant, complete and  terminating. In other words, we
have a decision procedure for determining the consistency of terms in
our extended feature logic.

For the purpose of showing {\em invariance} of our rules we distinguish
between {\em deterministic} and {\em non-deterministic} rules. Amongst
all our rules only rule (\ref{rule:SDis}) given in figure
\ref{fig:SRules2} and rule (\ref{rule:UnionDown}) are
non-deterministic while all the other rules are deterministic.
\begin{theorem}[Invariance]
\begin{enumerate}
\item If a decomposition rule transforms $C_{s}$ to $C'_{s}$ then $C_{s}$
  is consistent iff $C'_{s}$ is consistent.
\item Let $\I, \alpha$ be any interpretation, assignment pair and let 
  $C_{s}$ be any constraint system.
  \begin{itemize}
  \item If a deterministic simplification rule transforms  $C_{s}$ to
    $C'_{s}$ then:\\
    $\I, \alpha \models C_{s}$ iff $\I, \alpha \models C'_{s}$
  \item If a non-deterministic simplification rule applies to $C_{s}$
    then there is at least one non-deterministic choice which transforms
    $C_{s}$ to $C'_{s}$ such that:\\
    $\I, \alpha \models C_{s}$ iff $\I, \alpha \models C'_{s}$
  \end{itemize}
\end{enumerate}
\end{theorem}
A constraint system $C_{s}$ is in {\bf normal form} if no rules are
applicable to $C_{s}$.

Let $succ(x, f)$ denote the set:
\begin{itemize}
\item[] $succ(x, f) = \{ y \mid C_{s} \vdash x = \exists f: y \}$
\end{itemize}
A constraint system $C_{s}$ in normal form contains a {\bf clash} if
there exists a variable $x$ in $C_{s}$ such that {\em any} of the
following conditions are satisfied :
\begin{enumerate}
  \item $C_{s} \vdash x = a_{1}$ and 
    $C_{s} \vdash x = a_{2}$ 
       such that  $a_{1} \not\equiv a_{2}$
  \item $C_{s} \vdash x = c_{1}$ and 
    $C_{s} \vdash x = c_{2}$ 
       such that  $c_{1} \not\equiv c_{2}$
  \item $C_{s} \vdash x = \overline{S}$ and 
    $C_{s} \vdash x = \neg \overline{S}$\\
    where $\overline{S}$ ranges over $x,a, c, C$.
  \item $C_{s} \vdash x = \exists f: y$ and  $C_{s} \vdash x = a$
  \item $C_{s} \vdash f(x) \neq g(y)$ and 
    $succ(x, f) \cap succ(y,g) \neq \emptyset$
  \item $C_{s} \vdash x = f:\{x_{1}, \ldots , x_{n}\}_{=}$ and
    $\card{succ(x, f)} < n$
\end{enumerate}
If $C_{s}$ does not contain a clash then $C_{s}$ is called {\bf clash-free}.

The constraint solving process can terminate as soon as a {\em
  clash-free} constraint system in normal form is found or alternatively
all the choice points are exhausted.

The purpose of the {\em clash} definition is highlighted in the {\em
  completeness} theorem given below.

For a constraint system $C_{s}$ in normal form an {\em equivalence
  relation} $\simeq$ on variables occurring in $C_{s}$ is defined as follows:
\begin{itemize}
\item[] $x \simeq y$ if $C_{s} \vdash x = y$
\end{itemize}
For a variable $x$ we represent its equivalence class by $[x]$.

\begin{theorem}[Completeness]
  A constraint system $C_{s}$ in normal form is consistent iff $C_{s}$ is
  clash-free.
\end{theorem}
{\em Proof Sketch}: For the first part, let $C_{s}$ be a constraint system
containing a clash then it is clear from the definition of clash that
there is no interpretation $\I$ and $\I$-assignment $\alpha$ which satisfies
$C_{s}$.

Let $C_{s}$ be a clash-free constraint system in normal form.

We shall construct an interpretation $\R = <\U^{R}, .^{R}>$ and a variable
assignment $\alpha$ such that $\R, \alpha \models C_{s}$.

Let $\U^{R} = \V \cup \At \cup \C$.

The assignment function $\alpha$ is defined as follows:
\begin{enumerate}
\item For every variable $x$ in $\V$ 
  \begin{enumerate}
  \item if $C_{s} \vdash x = a$ then $\alpha(x) = a$
  \item if the previous condition does not apply then
    $\alpha(x) = choose([x])$ where $choose([x])$ denotes a unique
    representative (chosen arbitrarily) from the equivalence class $[x]$.
  \end{enumerate}
\item For every constant $c$ in $\C$:
  \begin{enumerate}
  \item if $C_{s} \vdash x = c$ then $\alpha(c) = \alpha(x)$
  \item if $c$ is a constant such that the previous condition does not
    apply then $\alpha(c) = c$
  \end{enumerate}
\item For every primitive concept $C$ in $\PP$:
  \begin{itemize}
  \item[] $\alpha(C) = \{\alpha(x) \mid C_{s} \vdash x = c\}$
  \end{itemize}
   
\end{enumerate}
The interpretation function $.^{R}$ is defined as follows:
\begin{itemize}
\item $f^{R}(x) = succ(x, f)$
\item $a^{R} = a$
\end{itemize}
It can be shown by a case by case analysis that for every constraint $K$
in $C_{s}$:\\ 
$\R, \alpha \models K$. 

Hence we have the theorem.

\begin{theorem}[Termination]
The~consistency checking procedure terminates in a finite number of
steps.
\end{theorem}
{\em Proof Sketch}: Termination is obvious if we observe the following
properties:
\begin{enumerate}
\item Since decomposition rules breakdown terms into smaller ones these
  rules must terminate.

\item None of the simplification rules introduce new variables and hence
  there is an upper bound on the number of variables.

\item Every simplification rule does either of the following:
  \begin{enumerate}
  \item reduces the `effective' number of variables.
 
    A variable $x$ is considered to be {\em ineffective} if it occurs
   only once in $C_{s}$ within the constraint $x = y$ such that rule
   (\ref{rule:SEquals}) does not apply. A variable that is not 
   {\em ineffective} is considered to be {\em effective}.

  \item adds a constraint of the form $x = \overline{C}$ where $C$ ranges
    over $y, a, c, C, \neg y, \neg a, \neg c, \neg C$ which means there
    is an upper bound on the number of constraints of the form $x =
    \overline{C}$ that the simplification rules can add.  This is so
    since the number of variables, atoms, constants and primitive
    concepts are bounded for every constraint system in basic form.

  \item increases the size of $succ(x, f)$. But the size of $succ(x, f)$
    is bounded by the number of variables in $C_{s}$ which remains
    constant during the application of the simplification rules. Hence
    our constraint solving rules cannot indefinitely increase the size of
    $succ(x, f)$.

 \end{enumerate}
\end{enumerate}

\section{NP-completeness}
In this section, we show that consistency checking of terms within the
logic described in this paper is NP-complete. This result holds even if
the terms involving set operations are excluded. We prove this result by
providing a polynomial time translation of the well-known NP-complete
problem of determining the satisfiability of propositional formulas
\cite{Garey:intractability}.
\begin{theorem}[NP-Completeness]
  Determining consistency of terms is NP-Complete.
\end{theorem}
{\em Proof:} Let $\phi$ be any given propositional formula
for which consistency is to be determined. We split our translation into
two intuitive parts : {\em truth assignment} denoted by $\Delta(\phi)$
and {\em evaluation} denoted by $\tau(\phi)$.

Let $a, b, \ldots $ be the set of propositional variables occurring in
$\phi$. We translate every propositional variable $a$ by a variable
$x_{a}$ in our logic. Let $f$ be some relation symbol. Let $true, false$
be two atoms.

Furthermore, let $x_{1}, x_{2}, \ldots$ be a finite set of variables
distinct from the ones introduced above.

We define the translation function $\Delta(\phi)$ by:
\begin{itemize}
\item[] $\begin{array}[t]{llll}
         \Delta(\phi) = & f: \{ true, false \} \sqcap \\
                        & \exists f:x_{a} \sqcap \exists f:x_{b}  \sqcap
                          \ldots \sqcap\\
                        & \exists f:x_{1} \sqcap \exists f:x_{2}  \sqcap
               \ldots
        \end{array}$
\end{itemize}
The above description forces each of the variable $x_{a}, x_{b}, \ldots$
and each of the variables $x_{1}, x_{2}, \ldots$ to be either equivalent
to {\em true} or {\em false}.

We define the evaluation function $\tau(\phi)$ by:
\begin{itemize}
\item[] $\tau(a) = x_{a}$
\item[] $\tau(S \& T) = \tau(S) \sqcap \tau(T)$
\item[] $\tau(S \vee T) = 
  x_{i} \sqcap \exists f: (f:\{\tau(S), \tau(T)\} \sqcap \exists f: x_{i})$\\
  where $x_{i} \in \{ x_{1}, x_{2}, \ldots \}$ is a new variable
\item[] $\tau(\neg S) =
   x_{i} \sqcap \exists f: (\tau(S) \sqcap \neg x_{i})$\\
  where $x_{i} \in \{ x_{1}, x_{2}, \ldots \}$ is a new variable
\end{itemize}
Intuitively speaking $\tau$ can be understood as follows.
Evaluation of a propositional variable is just its value; evaluating a
conjunction amounts to evaluating each of the conjuncts; evaluating a
disjunction amounts to evaluating either of the disjuncts and finally
evaluating a negation involves choosing something other than the value of
the term.

Determining satisfiability of $\phi$ then amounts to determining the
consistency of the following term:
\begin{itemize}
\item[] $\exists f: \Delta(\phi) \sqcap \exists f: (true \sqcap \tau(\phi))$
\end{itemize}
Note that the term $true \sqcap \tau(\phi)$ forces the value of
$\tau(\phi)$ to be $true$.  This translation demonstrates that
determining consistency of terms is NP-hard.

On the other hand, every deterministic completion of our constraint
solving rules terminate in polynomial time since they do not generate new
variables and the number of new constraints are polynomially bounded.
This means determining consistency of terms is NP-easy.
Hence, we conclude that determining consistency of terms is NP-complete.

\section{Translation to Sch\"onfinkel-Bernays class}
The Sch\"onfinkel-Bernays class (see \cite{Lewis:complexity}) consists of
function-free first-order formulae which have the form:
\begin{itemize}
\item[] $\exists x_{1} \ldots x_{n} \forall y_{1} \ldots y_{m} \delta$
\end{itemize}
In this section we show that the attributive logic developed in this
paper can be encoded within the Sch\"onfinkel-Bernays subclass of
first-order formulae by extending the approach developed in
\cite{Johnson:features}.  However formulae such as
$\forall~f:~(\exists~f:~(\forall f:~T))$ which involve an embedded
existential quantification cannot be translated into the
Sch\"onfinkel-Bernays class. This means that an unrestricted variant of
our logic which does not restrict the universal role quantification
cannot be expressed within the Sch\"onfinkel-Bernays class.

In order to put things more concretely, we provide a translation of every
construct in our logic into the Sch\"onfinkel-Bernays class.

Let $T$ be any extended feature term. Let $x$ be a variable {\em free} in
$T$. Then $T$ is consistent {\em iff} the formula $(x = T)^{\delta}$ is
consistent where $\delta$ is a translation function from our extended
feature logic into the Sch\"onfinkel-Bernays class. Here we provide only
the essential definitions of $\delta$:
\begin{itemize}
\item $(x = a)^{\delta} = x = a$
\item $(x = \neg a)^{\delta} = x \neq a$
\item $(x = f:T)^{\delta} =$\\
\hspace*{2ex}$f(x,y) \ \&\ (y = T)^{\delta} \ \&\ \forall y' (f(x, y') \rightarrow y = y')$\\
   where $y$ is a new variable

\item $(x = \exists f:T)^{\delta} = 
   f(x,y) \ \&\ (y = T)^{\delta}$\\
   where $y$ is a new variable

\item $(x = \forall f: a)^{\delta} = 
   \forall y (f(x, y) \rightarrow y = a)$

\item $(x = \forall f: \neg a)^{\delta} = 
   \forall y (f(x, y) \rightarrow  y \neq a)$

\item $(x = f: \{T_{1}, \ldots, T_{n}\})^{\delta} = $\\
\hspace*{2ex}$\begin{array}{llll}
     f(x, x_{1}) \ \&\ \ldots \ \&\ f(x, x_{n}) \&\ \\
     \forall y (f(x, y) \rightarrow  
       y = x_{1} \vee \ldots \vee y = x_{n}) \& \\
     (x_{1} = T_{1})^{\delta} \ \&\  \ldots \ \&\  (x_{1} = T_{n})^{\delta}
  \end{array}$\\
 where $x_{1}, \ldots, x_{n}$ are new variables

\item $(x = f: g(y) \cup h(z))^{\delta} =$\\
\hspace*{2ex}$\begin{array}{llll}
     \forall x_{i} (f(x,x_{i}) \rightarrow g(y,x_{i}) \vee h(z,x_{i})) \ \&\ \\
     \forall y_{i} (g(y,y_{i}) \rightarrow f(x,y_{i})) \ \&\ \\
     \forall z_{i} (h(z,z_{i}) \rightarrow f(x,z_{i}))
  \end{array}$

\item $(x = f:(y) \neq g(z))^{\delta} =$\\
\hspace*{2ex}$\forall y_{i} z_{j}
    (f(y,y_{i}) \ \&\  g(z,z_{i}) \rightarrow y_{i} \neq z_{i})$

\item $(x = S \sqcap T)^{\delta} = (x = S)^{\delta} \ \&\  (x = T)^{\delta}$
\end{itemize}
These translation rules essentially mimic the decomposition rules given
in figure \ref{fig:DRules}.

Furthermore for every atom $a$ and every feature $f$ in $T$ we need the
following axiom:
\begin{itemize}
\item $\forall a x (\neg f(a,x))$
\end{itemize}
For every distinct atoms $a, b$ in $T$ we need the axiom:
\begin{itemize}
\item $a \neq b$
\end{itemize}
Taking into account the NP-completeness result established earlier this
translation identifies a NP-complete subclass of formulae within the
Sch\"onfinkel-Bernays class which is suited for NL applications.


\section{Related Work}
Feature logics and concept languages such as KL-ONE are closely related
family of languages \cite{Nebel:attributive}. The principal difference
being that feature logics interpret attributive labels as functional
binary relations while concept languages interpret them as just binary
relations. However the integration of concept languages with feature
logics has been problematic due to the fact the while path equations do
not lead to increased computational complexity in feature logic the
addition of role-value-maps (which are the relational analog of path
equations) in concept languages causes undecidability
\cite{Schauss:undecidable}. This blocks a straightforward integration of
a variable-free concept language such as ALC \cite{Schauss:attributive}
with a variable-free feature logic \cite{Smolka:logic}.

In \cite{Manandhar:thesis} the addition of variables, feature symbols and
set descriptions to ALC is investigated providing an alternative method
for integrating concept languages and feature logics. It is shown that
set descriptions can be translated into the so called ``number
restrictions'' available within concept languages such as BACK
\cite{vonLuck:anatomy}. However, the propositionally complete languages ALV
and ALS investigated in \cite{Manandhar:thesis} are PSPACE-hard languages
which do not support set operations.

The work described in this paper describes yet another unexplored
dimension for concept languages - that of a restricted concept language
with variables, feature symbols, set descriptions and set operations for
which the consistency checking problem is within the complexity class NP.

\section{Summary and Conclusions}
In this paper we have provided an extended feature logic (excluding
disjunctions and negations) with a range of constraints involving set
descriptions. These constraints are set descriptions, fixed cardinality
set descriptions, set-membership constraints, restricted universal role
quantifications, set union, set intersection, subset and disjointness. We
have given a model theoretic semantics to our extended logic which shows
that a simple and elegant formalisation of set descriptions is possible
if we add relational attributes to our logic as opposed to just
functional attributes available in feature logic.

For realistic implementation of the logic described in this paper,
further investigation is needed to develop concrete algorithms that are
reasonably efficient in the average case.  The consistency checking
procedure described in this paper abstracts away from algorithmic
considerations and clearly modest improvements to the basic algorithm
suggested in this paper are feasible. However, a report on such
improvements is beyond the scope of this paper.

For applications within constraint based grammar formalisms such as HPSG,
minimally a type system \cite{Carpenter:typed} and/or a Horn-like
extension \cite{Hohfeld:relations} will be necessary.

We believe that the logic described in this paper provides both a
better picture of the formal aspects of current constraint based
grammar formalisms which employ set descriptions and at the same time
gives a basis for building knowledge representation tools in order to
support grammar development within these formalisms.

\section{Acknowledgments}
The work described here has been carried out as part of the EC-funded
project LRE-61-061 RGR (Reusability of Grammatical Resources). A longer
version of the paper is available in \cite{RGR:DeliverableB}. The work
described is a further development of the author's PhD thesis carried out
at the Department of Artificial Intelligence, University of Edinburgh. I
thank my supervisors Chris Mellish and Alan Smaill for their guidance.  I
have also benefited from comments by an anonymous reviewer and
discussions with Chris Brew, Bob Carpenter, Jochen D\"orre and Herbert
Ruessink.

The Human Communication Research Centre (HCRC) is supported by the
Economic and Social Research Council (UK).

\bibliographystyle{named}

\end{document}